# Shape Programmable Magnetic Pixel Soft Robot[*]

Ran Zhao[1,2,*], Hanchen Yao[1], Houde Dai[1,*]

[1]Quanzhou Institute of Equipement Manufacturing of Haixi Institutes, Chinese Academy of Sciences, Quanzhou 262000, China

[2]Zhongyuan-Petersburg Aviation College, Zhongyuan University of Technology, Zhengzhou 450000, China

[*]E-mail: dhd@fjirsm.ac.cn; zhaoran@zut.edu.cn

**Abstract:** Magnetic response soft robot realizes programmable shape regulation with the help of magnetic field and produces various actions. The shape control of magnetic soft robot is based on the magnetic anisotropy caused by the orderly distribution of magnetic particles in the elastic matrix. In the previous technologies, magnetic programming is coupled with the manufacturing process, and the orientation of magnetic particles cannot be modified, which brings restrictions to the design and use of magnetic soft robot. This paper presents a magnetic pixel robot with shape programmable function. By encapsulating NdFeB/gallium composites into silicone shell, a thermo-magnetic response functional film with lattice structure are fabricated. Basing on thermal-assisted magnetization technique, we realized the discrete magnetization region distribution on the film. Therefore, we proposed a magnetic coding technique to realize the mathematical response action design of software robot. Using these methods, we prepared several magnetic soft robots based on origami structure. The experiments show that the behavior mode of robot can be flexibly and repeatedly regulated by magnetic encoding technique. This work provides a basis for the programmed shape regulation and motion design of soft robot.

# 1 Introduction

Magnetic soft robot, as a kind of untethered robot, can implement tasks as cell manipulation, medical image acquisition, drug delivery and non- invasive intervention. Comparing to light/thermal, chemical or electrical actuated soft robots[1-4], magnetic soft robots have the advantages of fast response, unlimited endurance and no obstruction restrictions [5] .Therefore, it has made great progress in re- cent years.

The motion of magnetic soft robot comes from the response of magnetic particles wrapped in flex- ible matrix when applying a magnetic field.These particles can be soft magnetic materials (Fe, Ni and $Fe_3O_4$)[6,7] or magnetic hard magnetic materials ($CrO_2$ and NdFeB) [8-10]. The soft robot based on hard magnetic materials has high residual magnetization. Its programmable shape control can be realized by con- figuring the magnetic anisotropy, which called pro- grammable magnetization technique. However, the magnetizing process is always coupled with the robot manufacturing process, and the magnetic anisotropy distribution is unchangable. The introduction of phase-transition polymer has changed this defect (11,12). By heating, magnetic particles can become free domains, be re- oriented, and be locked in the cooling state. But the heating temperature required is too high for biomedical applications.

So far, magnetic programming is usually implemented in continuous media [13,14], and there is no obvious boundary between different magnetized re-film can be observed by the CMOS camera.The proposed themo-magnetic response functional film is made of NdFeB/galium composites, coating with Silicone. As given in Fig.1b, the film has a discrete lattice structure, each basic cell is called magnetic pixel. In each magnetic pixel, the NdFeB microparticles are wrapped and uniformly distributed in the liquid-metal matrix regions. Some composites with lattice structure are proposed, realizing the independent programming of discrete magnetization region. This makes it possible for more refined morphological control for soft robot. Cui proposed the concept of magnetic encod- ing一a technique to generate magnetic response by coding magnetic vectors on discrete magnetic units [15]. However, this technique can only deal with vertical vector coding.

In summery, there is still a lack of sufficient tech- nology to realize repeatable magnetization at low heating temperature and arbitrary magnetic vectors encoding. This paper presents a magnetic pixel robot by encapsulating NdFeB/gallium composites into silicone shell. We developed a thermal-assisted 3D magnetization technique,to realize the discrete magnetization region distribution on the film.We proposed a magnetic encoding technique to realize the mathematical response action design for soft robot. The details are given in follow sections.

## 2. Principle

### 2.1 Reprogrammable 3D magnetization

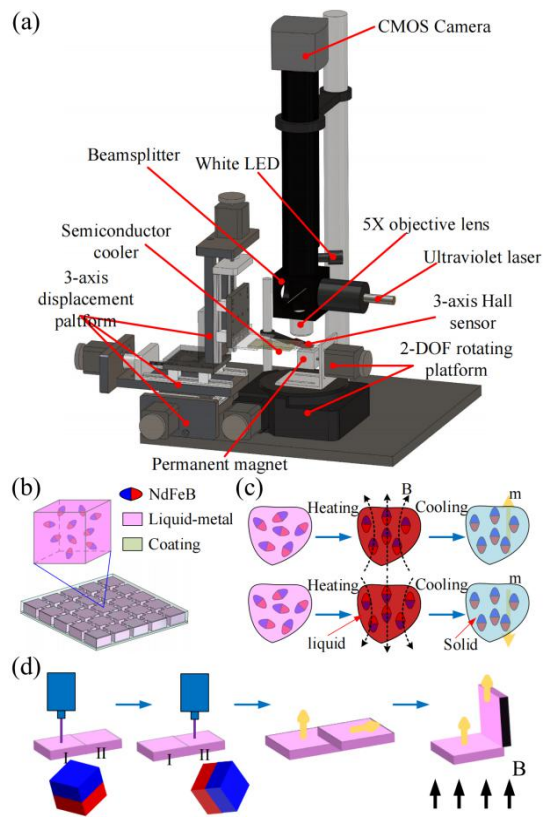

**Fig. 1** System design and mechanism of reprogramable 3D magnetization in MFs-LM films:(a)3D magnetic vector programming system, (b)Structures of MFs-LM film and its magnetic pixel,(c)Working mechanism of MFs-LM composites and (d)Programming a magnetic anisotropy in the MFs-LM film and generating a magnetic response action.

Fig.1a shows the physical apparatus for patterning 3D discrete magnetization in NdFeB/Ga/Silicone composite films. The system consists of a 3-axis positioning platform, a 2-DOF rotating platform, a 3-axis Hall sensor, a square magnet (N52,

24×24×24 mm³, surface magnetic intensity of 735 mT), a semiconductor cooler, and a coaxial optical system (including a 405 nm ultraviolet laser and a CMOS camera).Based on the 5-axis motion platform, a 3D vector magnetic field of any direction and strength can be generated.a 3D vector magnetic field of any direction and strength can be generated. A 10W, 800 μm resolution UV laser is used to heat MFs- LM composites.. The semiconductor cooler is used to reduce the temperature of the non heated region of the film. At last, the change of the functional film can be observed by the CMOS camera.The proposed thermal-magnetic response functional film is made of NdFeB/galium composites, coating with Silicone.

As given in Fig.1b, the film has a discrete lattice structure, each basic cell is called magnetic pixel. In each magnetic pixel, the NdFeB microparticles are wrapped and uniformly distributed in the liquidmetal matrix. Fig.1c exhibits the mechanism of reprogramable magnetization based on the MFs-LM composites.When heating a certern region of the materials by the laser, gallium can be transformed from solid to liquid phase.Then NdFeB particles will be reoriented under the programming magnetic field, and form macro magnetic anisotropy in the heated region. Here, the required heating temperature is only 40 ◦ C. When the magnetization process is completed, the laser stops working, and the liquid-metal is converted into solid phase again under the action of semiconductor cooler. By repeating this process, we can program different magnetic anisotropy in the film. In Fig.1d, we shows how to program magnetic anisotropy in different regions of the film and produce magnetic response actions. By programming on region I and region II of the film respectively, the residual magnetization with vertical and horizontal directions is obtained.Finally, under the exiting of vertical magnetic field, a simple bending action is generated.

**2.2 Magnetic encoding**

Here,we proposed a mathematical description of 3D magnetic vector encoding, a vector matrix **A** is used to describe the magnetization vector direction for each magnetic pixel, which is given as,

$$A = \begin{bmatrix} a_{11} & \cdots & a_{1n} \\ \vdots & \ddots & \vdots \\ a_{m1} & \cdots & a_{mn} \end{bmatrix}_{m \times n}$$

where $a_{ij}=(\alpha_{ij}, \beta_{ij}, \gamma_{ij})^T$, represents the 3D magnetization vector of the *ij*-th magnetic pixel. As shown in Fig2.a, $\alpha_{ij}$, $\beta_{ij}$ and $\gamma_{ij}$ are the angles between magnetization vector and *x*, *y* and *z* axis, respectively.

Fig.2b shows how to design the profile of the soft robot on the magnetic pixel array, the part of the matrix with a value of 0 represents the area to be removed.Then, the required response action can be obtained by magnetic coding for each pixel in the reserved area. Finally, according to equation (1), the following magnetization vevtor matrix **M** can be obtained,

$$\mathbf{M} = \mathbf{m} \cdot \mathbf{A} \tag{2}$$

where *m* is the remanent magnetization.

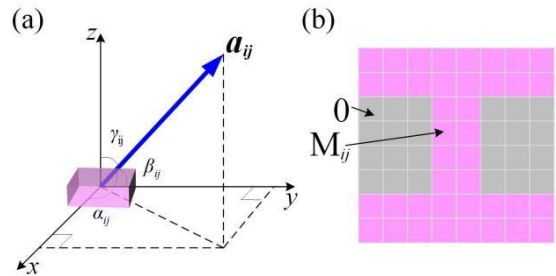

**Fig. 2** Illustration of magnetic encoding: (a) Magnetic vector of the *ij*-th magnetic pixel, (b) Contouring and encoding on the magnetic pixel array.

## 3 Material

### 3.1 Preparation of MFs-LM composites

Here, the ferromagnetic liquid-metal functional composites is prepared by homogenously mixing LM of gallium (the melting point of 29.6°C, purity of 99.9 %) and ferromagnetic neodymium–iron–boron (NdFeB) microparticles.The average particle size of NdFeB particle (Xinnuode Transmission Devices, Guangzhou) was 6.0 μm³. As shown in Fig.2a, the unmagnetized NdFeB powder and liquid gallium, with a volume ratio of 4:6, were mixed by a mechanical agitator.The mixture was heated to

50°C and stirred in air enviroment for 5 min. The NdFeB/Ga composites was magnetized in a pulsed magnetic field with magnetic strength of 2 T. The magnetized MFs-LM shows high rheological and shape reconfigurable properties, which is similar with the previously reported other LM composites [11-13].

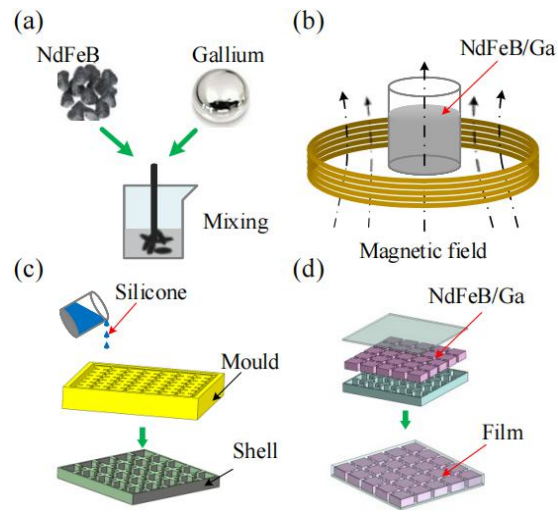

**Fig. 3** Fabrication process:(a) Mixing and (b)Magnetizing of MFs-LM composites,(c) preparing and (d)encapsulating the MFs-LM composites into the silicone shell.

### 3.2 Fabrication of magnetic pixel film

In Fig.2c, a biocompatible material — Silicone (Ecoflex00-50, Smooth-On, America) was used for coating the FMs-LM composites. And a 3D-printed mold was used to fabricate the shell of magnetic pixel film.The designed size of each magnetic pixel is 2×2 mm$^2$.The Silicone solvent was poured into the mold,and cured in vacuum for 1 h, obtaining a shell with thickness of 100 μm.As shown in Fig.2d, the MFs-LM plasticine with thickness of 600 μm was manually encapsulated into the silicone shell. The total thickness of the functional film is 800 μm.

## 4 Experiments and Discussion
### 4.1 Experimental results

In experiments, we tested the performance of the magnetic pixel film,and manufactured several magnetic soft robots with different structures. By configuring different magnetization regions on the soft robot, we showed that the soft robot can perform different motions, the results are given in Fig.4. Figs.4a and 4b exhibit the graphs of the proposed magnetic pixel film, and the magnetic imaging of encoding letters "CAS" in the film.In Figs.4c - 4h, we fabricated there robots with I, Y and ring shapes, respectively.When implementing different magnetic encoding schemes,they can perform different response actions.

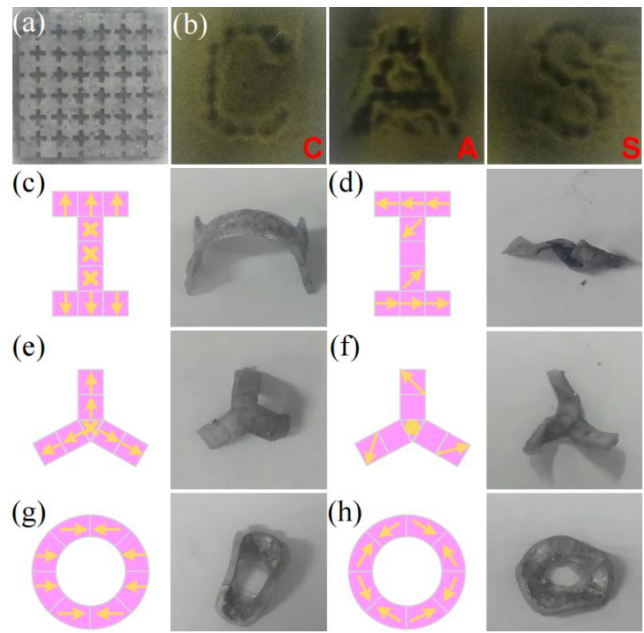

**Fig. 4** Results of magnetic encoding in the film:(a)Photograph of magntic pixel film,(b)Magnetic imaging of coding letters, (c)Bending and (d)Twisting of I shape robot, (e)Standing and (f) Torsion of Y shape robot,(g) Folding and (h) Higher-order bending of ring shape robot.

Another case is shown in Fig.5, a flexible zaxis positioning plateform was built and tested.The plateform has a square flat and four supporting arms,which can move along z-axis direction when applying a vertical magnetic field (shown in Figs.5a and 5b).The encoding scheme is given in Fig.5c. The positioning curve is shown in Fig.5d, the maximum displacement is 3.1 mm under the magnetic field of 54.5 mT.

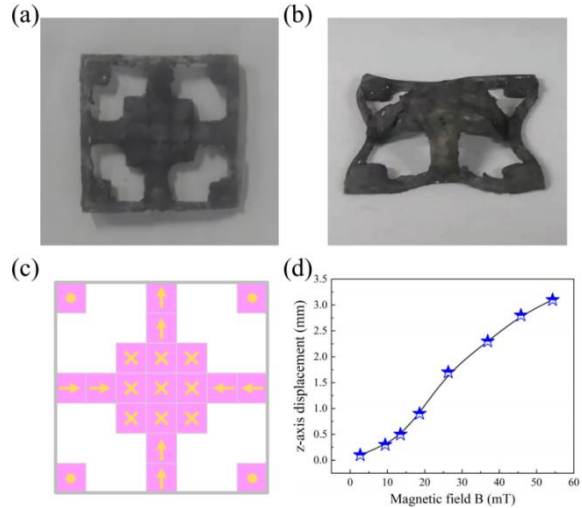

**Fig. 5** Flexible z-axis platform:(a)Photograph of z-axis platform,(b)Moving along z-axis direction when applying a magnetic field, (c)Magnetic encoding and (d) Displacement of z-axis.

Fig.6 shows the magnetic respones of the crossshape soft robot,when implemented three different magnetic encodes.As shown in Figs.6a and 6b, the robot performs as a fan with four blades. In Fig.6c,when using another magnetic encodes, the robot can stand up (Fig.6d).This robot also can work as a magnetic graspper or capsule, capturing or transport a target(given in Figs.6e and 6f).

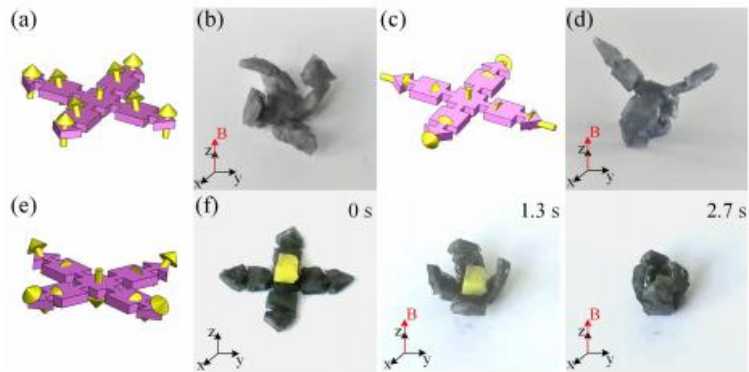

**Fig. 6** Cross-shape robot:(a)Encode 1,(b)Fan, (c)Encode 2, (d)Standing,(e)Encode 3, and (f)grasping.

Fig.7 exhibits a magnetic origami robot, consisting of eight triangular plates. Four kinds of encodes (as shown in Figs.7a, 7c, 7e, and 7g) were designed to generated different shapes of parallel folding,diagonal folding, pyramid and diagonal double folding (given in Figs.7b, 7d, 7f, and 7h). The cases given in Figs.4, 5, 6 and 7 indicate that through the repeatable magnetization techniques, a robot can be re-encoded to generate different response actions. And this ability of reprogrammable shape helps the robot to perform different tasks. Furthermore, with the reconfigurable function, the reusability of soft robot is greatly improved.

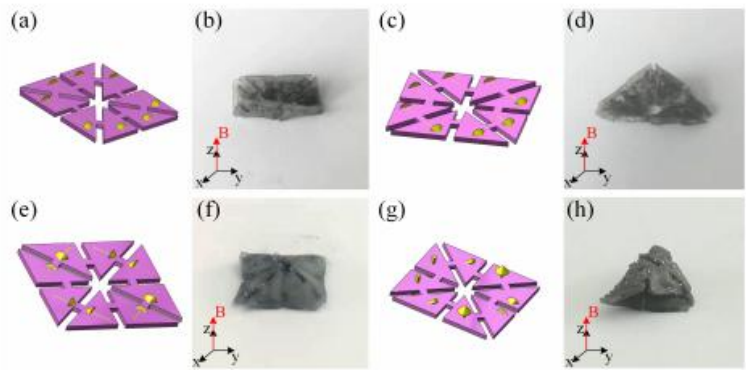

Fig. 7 Magnetic origami robot:(a)Encode 1,(b)Parallel folding, (c)Encode 2, (d) Diagonal folding, (e)Encode 3, (f)Pyramid, (g)Encode 4, and (h)Diagonal double folding.

**4.2 Comparison of exiting techniques**

In this section, we analyzed and compared the reported magnetic programming techniques. The details are shown in Table 1. Our technique enables repeatable 3D magnetization, requiring lower heating temperature. And we provides a mathematical model of magnetic vector encoding.

Table 1 Comparison of existing magnetic programming techniques

| Ref | Materials | Fabrication methods | Medium | Magnetic programming technique |
|---|---|---|---|---|
| 16 | NdFeB/UV resin | UV lithography | Continue | Unrepeatable 3D magnetization heating-free |
| 17 | $CrO_2$/PDMS | Template-aided | Continue | Repeatable 3D magnetization, 110°C heating temperature, |
| 18 | NdFeB/PCL/Silicone | Template-aided | Continue | repeatable 2D magnetization, 80 °C heating temperature |
| 15 | Nano Fe/PMMA | UV lithography | Discreet | Unrepeatable 3D magnetization magnetic encoding |
| This work | NdFeB/Ga/Silicone | Template-aided | Discreet | Repeatable 3D magnetization, magnetic vector encoding, 40°C heating temperature |

**5 Conclusion**

In this work, we presented a shape programmable magnetic pixel soft robot realizes. The proposed robot is made NdFeB/Ga/silicone composites, and has a lattice structure. The programmable magnetization and the equipment are developed to implement Repeatable magnetization in the magnetic pixel film. A mathematical model of magnetic encoding was also proposed. The experimental results have testified our concept. The contributions of this paper are as follows:

i. We designed a magnetic pixel film, which has thermal/magnetic response function and can be repeatedly magnetized.

ii. We developed a 3D magnetic vector programming equipment, which can configure the magnetic anisotropy in any direction to independent magnetic pixel.

iii. We presented a technique of magnetic encoding, and presented its mathematical model, which provides a basis for automatic robot behavior design.

The next stage of our work is to reduce the size of magnetic pixels, and to manufacture milli-/micro scale magnetic soft robot.

**Compliance with ethics guidelines**

Ran ZHAO and Hou-de DAI declare that they have no conflict of interest.